\documentclass{article}
\usepackage{spconf,amsmath,graphicx,hyperref}
\usepackage{listings}
\usepackage{float}
\usepackage{multirow}
\usepackage{booktabs}

\title{Structured Information for Improving Spatial Relationships in Text-to-Image Generation}
\name{Sander Schildermans$^{\dagger}$, Chang Tian$^{\dagger}$$^{*}$\thanks{* Chang Tian and Ying Jiao are co-first authors.}, Ying Jiao$^{\dagger}$$^{*}$\thanks{Corresponding authors: Chang Tian (namechangtian@163.com) and Ying Jiao (ying.jiao@kuleuven.be).}, Marie-Francine Moens$^{\dagger}$ }
\address{$^{\dagger}$ KU Leuven, Leuven, Belgium \\
\texttt{\{sander.schildermans\}@student.kuleuven.be}}
\begin{document}
%
\maketitle
\begin{abstract}
Text-to-image (T2I) generation has advanced rapidly, yet faithfully capturing spatial relationships described in natural language prompts remains a major challenge. Prior efforts have addressed this issue through prompt optimization, spatially grounded generation, and semantic refinement. This work introduces a lightweight approach that augments prompts with tuple-based structured information, using a fine-tuned language model for automatic conversion and seamless integration into T2I pipelines. Experimental results demonstrate substantial improvements in spatial accuracy, without compromising overall image quality as measured by Inception Score. Furthermore, the automatically generated tuples exhibit quality comparable to human-crafted tuples. This structured information provides a practical and portable solution to enhance spatial relationships in T2I generation, addressing a key limitation of current large-scale generative systems.
\end{abstract}
\section{Introduction}
\label{sec:intro}
Recent advances in text-to-image generation have been driven by diffusion-based models such as Stable Diffusion~\cite{LDM} and DALL-E 3~\cite{dall-e2}. These systems are able to synthesize highly realistic and creative imagery from natural language descriptions, with applications in various sectors such as art, advertising and the entertainment industry~\cite{Promptcharm}. Despite these achievements, faithfully representing spatial relationships described in natural language prompts remains a major challenge for state-of-the-art models.

Previous attempts have explored approaches such as prompt optimization~\cite{promptist, Promptcharm, promptify}, spatially grounded generation~\cite{reco, spatext, densediffusion}, and semantic refinement~\cite{pickapic,dpt}. However, these approaches often incur substantial computational overhead and face challenges in generalization due to their complex designs.

In this paper, we introduce a \textbf{lightweight, plug-and-play solution} that augments plain natural language prompts with structured, tuple-based information of objects and their spatial relationships.
\begin{figure*}[t]
    \centering
\includegraphics[width=0.86\textwidth]{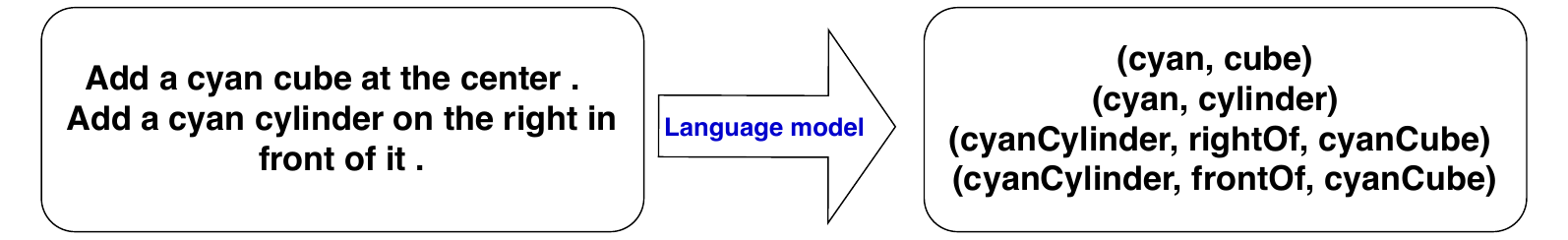}
    \caption{An example of structured information extraction. Given a plain natural language prompt, the fine-tuned language model produces tuple-based representations that explicitly encode objects and their spatial relations, providing a compact and model-friendly input format.}
    \label{fig:tuples_final}
\end{figure*}
Concretely, a fine-tuned T5-small~\cite{t5-small} model automatically converts natural language texts into the structured information as shown in Figure~\ref{fig:tuples_final}. 
The structured information is appended to the plain natural language prompts and directly fed into Stable Diffusion XL~\cite{sdxl} (SDXL), without any modification to its architecture or parameters.

Our main contributions are:
\begin{itemize}
    \item We propose a novel tuple-based representation that explicitly encodes objects and their spatial relations in a compact, model-friendly format.
    \item We propose a lightweight and portable converter that transforms natural language into structured information, built on T5-small with only 60M parameters and requiring limited training data.
    \item Extensive experiments demonstrate improved handling of spatial relationships, along with higher image quality as measured by Inception Score. Our method outperforms plain prompt inputs and two leading baselines, RealCompo~\cite{realcompo} and DPT-T2I~\cite{dpt}.
    \item We propose a plug-and-play generation pipeline that enhances spatial alignment while remaining portable and easy to integrate.
\end{itemize}
\section{Related Work}
\label{sec:related_work}
Relevant studies can be grouped into text-to-image generation and semantic parsing. However, tuple extraction has not yet been applied to automatic prompt construction for text-to-image generation. Distinct from prior works, our method introduces the idea to enhance the representation of spatial relationships in text-to-image generation.
\subsection{Text-to-image Generation}
Deep learning has significantly advanced numerous applications~\cite{paint4poem,antiover,fightingagainst,2024generic,2025using,tian2025deep,tian2025large}, with text-to-image generation representing an important area of progress.
Efforts to enhance the performance of text-to-image generation can be broadly classified into following research directions.

\textbf{Prompt Optimization.} Prior studies focus on improving the clarity and aesthetics of prompts. Promptist~\cite{promptist} applies reinforcement learning to rewrite prompts for greater coherence. Promptify~\cite{promptify} leverages large language models to generate stylistic variations or prompt extensions. PromptCharm~\cite{Promptcharm} enables interactive refinement of prompts. While these approaches enhance image quality, they provide limited gains in the spatial relationship~\cite{realcompo,dpt}.

\textbf{Layout-Aware and Spatially Grounded Generation.}
Recent works aim to enhance spatial grounding and layout control in text-to-image generation. ReCo~\cite{reco} introduces position tokens to improve spatial alignment, while SpaText~\cite{spatext} integrates segmentation maps with region-specific text. DenseDiffusion~\cite{densediffusion} modifies intermediate attention maps for more precise layout control. VPGen~\cite{VPGen} adopts a modular pipeline that sequentially generates objects, layouts, and renderings. RealCompo~\cite{realcompo} offers a training-free approach that combines diffusion and layout-aware models through a noise-balancing mechanism, improving object placement while maintaining realism. Despite their effectiveness, these methods often involve substantial computational overhead.

\textbf{Cross-Modal Alignment and Semantic Refinement.}
ELLA~\cite{ella} integrates LLM representations at different timesteps into the diffusion process to refine semantic consistency. 
Feedback-based methods further enhance alignment: Pick-a-pic~\cite{pickapic} and ImageReward~\cite{imagereward} employ human preference modeling, while DreamSync~\cite{dreamsync} applies self-training with vision–language models. DPT-T2I~\cite{dpt} introduces a discriminative adapter for T2I models, probing their discriminative capabilities and leveraging fine-tuning to improve text–image alignment.
\subsection{Semantic Parsing}
Prior research in natural language processing has demonstrated the effectiveness of converting natural language into structured, machine-readable formats~\cite{shicommands, liu2025survey}. For example, \cite{tuple_extraction} introduced semantic tuple extraction for text summarization, while \cite{survey} provided a broader overview of parsing methods across diverse applications.
\section{Method}
\label{sec:method}
\begin{figure*}[t]
    \centering
    \includegraphics[width=0.87\textwidth]{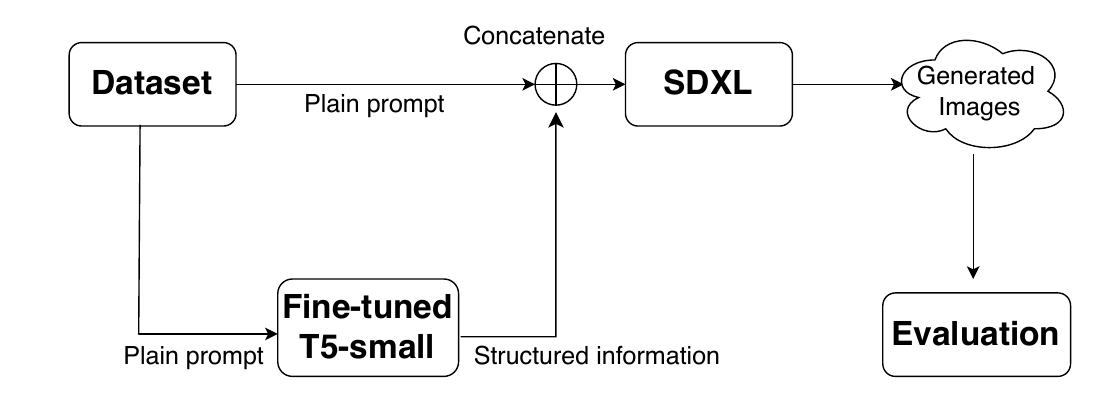}
    \caption{An overview of the StructuredPrompter framework.}
    \label{fig:pipeline}
\end{figure*}

We first construct the datasets used for fine-tuning T5-small and for evaluation (Section~\ref{sec:31}). Next, we fine-tune T5-small and employ it to generate structured information, which augments plain prompts for image generation with Stable Diffusion XL (Section~\ref{sec:32}). Finally, we evaluate the generated images (Section~\ref{sec:33}). A visualization of the proposed pipeline is presented in Figure~\ref{fig:pipeline}. We refer to our method as \textbf{StructuredPrompter}.

\subsection{Dataset Construction}
\label{sec:31}
We use the TV-Logic dataset~\cite{tvlogic}, which provides relational text descriptions paired with synthetic images of simple colored shapes. Its emphasis on spatial relations makes it particularly suitable for evaluating the proposed method. To reduce ambiguity and noise in evaluation, we restrict the dataset to two-object scenes, excluding the counting and manipulation subsets. The resulting subset used in our experiments consists of:
\begin{itemize}
\item 500 training prompts,
\item 100 validation prompts,
\item 1000 test prompts.
\end{itemize}

For each prompt sample in the training and validation sets, we generate tuple-based structured information with the assistance of GPT-4o. This structured representation consists of two components:
\begin{itemize}
    \item \textbf{Object tuples:} (color, shape),
    \item \textbf{Spatial relation tuples:} (subjectID, relation, objectID).
\end{itemize}
An example of the conversion from plain text to the tuple format is shown in Figure~\ref{fig:tuples_final}. This format is compact, efficiently generated, and well aligned with the natural language patterns that diffusion models are trained on.

For each prompt in the validation set, we manually extract structured information to serve as a reference for comparison with the structured information generated by the fine-tuned language model during empirical evaluation. In total, the validation set contains 100 prompts, each paired with manually extracted structured information.
\subsection{Augmented Prompts}
\label{sec:32}
Given a plain natural language prompt $X$ as input and the GPT-4o–extracted structured information $Y$ as ground truth, we fine-tune the language model $f(\cdot)$ to predict $\hat{Y}$. The fine-tuning process is optimized with cross-entropy loss, as shown in the following equations:

\begin{equation}
\hat{Y} = f(X),
\end{equation}

\begin{equation}
\mathcal{L}_{CE} = - \sum_{i=1}^{N} Y_i \log \hat{Y}_i,
\end{equation}

where $N$ is the number of tokens in the structured output, $Y_i$ is the ground-truth token distribution at position $i$, 
and $\hat{Y}_i$ is the predicted probability distribution over the vocabulary at position $i$.

The training process is evaluated on the validation set every five epochs. If performance does not improve within the subsequent five epochs, we consider the training to have converged. 

In this work, we employ T5-small as the language model for structured information extraction. On the validation set, it achieved scores above 0.98 in both BLEU and ROUGE, confirming its reliability. Its compact model size further demonstrates the feasibility of lightweight tuple extraction. Detailed statistics are reported in Table~\ref{tab:training-metrics}.

The extracted structured information is then concatenated with plain natural language prompts to form augmented prompts, which are provided as inputs to Stable Diffusion XL for image generation. This process requires no modifications to the diffusion model and adds minimal overhead, making it portable across text-to-image models.
\subsection{Evaluation}
\label{sec:33}

\textbf{Spatial Relationships.}  
We employ Qwen2.5-VL-3B-Instruct~\cite{qwen} as a judge to verify whether the generated images correctly convey the information specified in the plain prompts. For instance, given the prompt “B is on the right of A" and a generated image, we query Qwen2.5-VL-3B-Instruct with “Is B on the right of A? Please answer Yes or No." A correctly generated image should support a Yes response.

In addition to spatial relationships, we also evaluate color and shape alignment. The results are in the Table~\ref{tab:qwen-eval-results}.

\textbf{Image Quality.}  
We further assess the overall visual quality of the generated images using the Inception Score~\cite{is}.  
\begin{table*}[t]
\centering
\renewcommand{\arraystretch}{1.3}
\resizebox{0.9\textwidth}{!}{%
\begin{tabular}{|c|cc|cc|cc|}
\hline
\multirow{2}{*}{\textbf{\# Training Samples}} & \multicolumn{2}{c|}{\textbf{Llama-3.2-1B}} & \multicolumn{2}{c|}{\textbf{Llama-3-8B}} & \multicolumn{2}{c|}{\textbf{T5-small}} \\
\cline{2-7}
 & \textbf{BLEU (Epoch)} & \textbf{ROUGE (Epoch)} & \textbf{BLEU (Epoch)} & \textbf{ROUGE (Epoch)} & \textbf{BLEU (Epoch)} & \textbf{ROUGE (Epoch)} \\
\hline
100 & 0.0000 (0) & 0.08 (0) & 0.0306 (0) & 0.25 (0) & 0.0689 (20) & 0.21 (20) \\
200 & 0.0097 (10) & 0.08 (10) & 0.0367 (20) & 0.30 (20) & 0.9156 (20) & 0.894 (20) \\
300 & 0.0125 (20) & 0.18 (20) & 0.0493 (0) & 0.35 (0) & 0.9828 (20) & 0.99 (20) \\
400 & 0.0000 (0) & 0.08 (0) & 0.0544 (20) & 0.26 (20) & 0.9844 (20) & 0.99 (20) \\
\textbf{500} & 0.0113 (10) & 0.14 (10) & 0.0509 (20) & 0.28 (20) & \textbf{0.9889} (15) & \textbf{0.99} (15) \\
\hline
\end{tabular}%
}
\caption{Best BLEU and ROUGE scores on the validation set (with corresponding convergence epochs) for Llama-3.2-1B, Llama-3-8B and T5-small across different training sample sizes. We use T5-small, fine-tuned on 500 training samples, as our language model for structured information extraction.}
\label{tab:training-metrics}
\end{table*}
\section{Experiments}
\label{sec:experiments}
\subsection{Implementation Details}
We conduct experiments using three random seeds (40, 41, and 42). The training batch size is 4 with a gradient accumulation step of 8. The learning rate is set to 1e-4, optimized using Adam with 10 warm-up steps and a weight decay of 0.01. 
All experiments are implemented in PyTorch 2.5.1 and Transformers 4.50, and executed on an NVIDIA A100 GPU. Additional details are available in the GitHub repository of this work~\footnote{The code and data will be publicly available upon publication at https://github.com/Sander445/StructuredPrompter.}.
\begin{table*}[h]
\centering
\begin{tabular}{l l c c c}
\toprule
\textbf{Model} & \textbf{Input Type} & \textbf{Spatial} & \textbf{Color} & \textbf{Shape} \\
\midrule
SDXL & Plain & 0.446 $\pm$ 0.015 & 0.484 $\pm$ 0.021 & 0.470 $\pm$ 0.008 \\
SDXL & Plain + Structured (ours) & \textbf{0.473 $\pm$ 0.004} & \textbf{0.499 $\pm$ 0.012} & \textbf{0.483 $\pm$ 0.015} \\
DPT-T2I & Plain & 0.465 $\pm$ 0.024 & 0.452 $\pm$ 0.012 & 0.479 $\pm$ 0.021 \\
RealCompo & Plain & 0.310 $\pm$ 0.009 & 0.339 $\pm$ 0.021 & 0.375 $\pm$ 0.027 \\
\bottomrule
\end{tabular}
\caption{Evaluation results with Qwen2.5-VL-3B-Instruct on the test set (mean $\pm$ std over 3 seeds). Reported values indicate the proportion of “Yes” responses across all generated images. The experimental results are reported as averages over three runs with 3 random seeds.}
\label{tab:qwen-eval-results}
\end{table*}
\begin{table*}[h]
\centering
\label{tab:manual}
\begin{tabular}{l l c c c}
\toprule
\textbf{Model} & \textbf{Conversion Method} & \textbf{Spatial} & \textbf{Color} & \textbf{Shape} \\
\midrule
SDXL & Manual & 0.453 $\pm$ 0.047 & 0.473 $\pm$ 0.040 & 0.450 $\pm$ 0.035 \\
SDXL & Fine-tuned T5-small (ours) & 0.457 $\pm$ 0.031 & 0.480 $\pm$ 0.030 & 0.450 $\pm$ 0.026 \\
\bottomrule
\end{tabular}
\caption{Evaluation results of Qwen2.5-VL-3B-Instruct on curated 100 prompt samples for structured information extraction, comparing human annotations with language model outputs (mean $\pm$ std over 3 seeds).}
\end{table*}
\begin{table}[h]
\centering
\begin{tabular}{l l c}
\toprule
\textbf{Model} & \textbf{Input Type} & \textbf{Inception Score} \\
\midrule
SDXL        & Plain        & 6.71 $\pm$ 0.52 \\
SDXL        & Plain + Structured (ours)   & \textbf{6.94 $\pm$ 0.58} \\
DPT-T2I   & Plain        & 5.58 $\pm$ 0.43 \\
RealCompo     & Plain        & 6.90 $\pm$ 0.38 \\
\bottomrule
\end{tabular}
\label{tab:InScore}
\caption{Inception Score (mean $\pm$ std) for all models.}
\end{table}
\subsection{Impact Of Parameter Scale}
As shown in Table~\ref{tab:training-metrics}, we adopt fine-tuned T5-small as the backbone model for structured information extraction. Using 500 samples generated with GPT-4o, it achieves a BLEU score of 0.98 and a ROUGE score of 0.99. These results demonstrate that T5-small is a lightweight yet effective extractor of structured information. In contrast, Llama-3.2-1B and Llama-3-8B yield unreliable outputs, frequently producing malformed sequences. A plausible explanation is that the larger parameter scale of the Llama models leads to overfitting when fine-tuned with a limited number of samples.
\subsection{Quantitative Results}
As shown in Table~\ref{tab:qwen-eval-results}, appending structured information substantially improves the modeling of spatial relationships. This improvement arises because the structured representation emphasizes spatial elements and attributes while maintaining strong alignment with the natural language patterns on which diffusion models are trained. Notably, our method also outperforms plain prompts in terms of color and shape alignment between the generated images and the input prompts. These results demonstrate that tuple-based structured information provides effective features for text-to-image generation with SDXL. An  explanation is that the vast pre-training data of SDXL contains ontology-like knowledge, which facilitates integration with structured information.

Moreover, in Table~\ref{tab:qwen-eval-results}, our method outperforms two leading baselines in the community, DPT-T2I and RealCompo, demonstrating that it offers a portable, plug-and-play, and effective solution for enhancing spatial relationships in text-to-image tasks.

As shown in Table 4, we use the Inception Score (IS) to evaluate the visual quality of the generated images, where higher values indicate better quality. After appending structured information, our method achieves higher IS values than plain prompts, outperforming or matching two leading baselines. These results further confirm the effectiveness of structured information for text-to-image generation.
\subsection{Qualitative Results}
\begin{figure}[th]
    \centering
    \begin{tabular}{c c}
        \includegraphics[width=0.38\columnwidth]{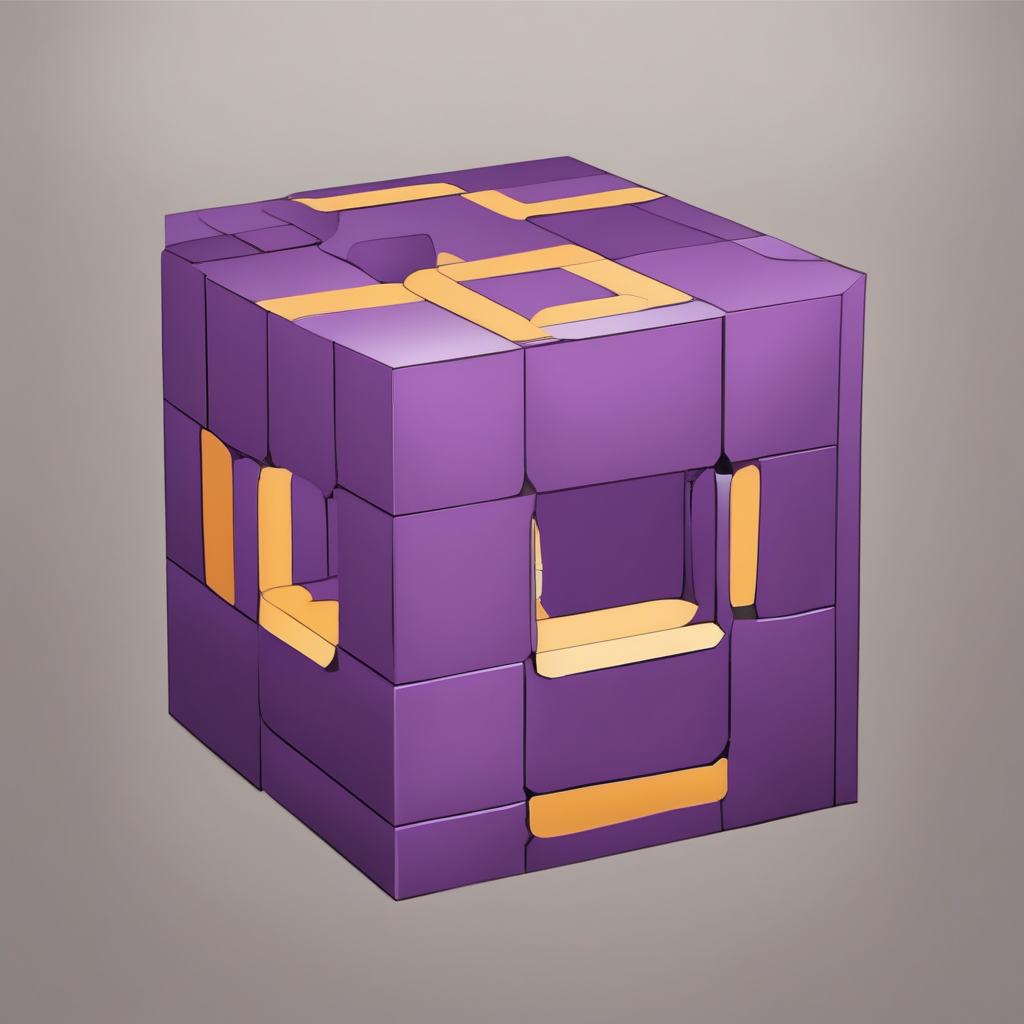} &
        \includegraphics[width=0.38\columnwidth]{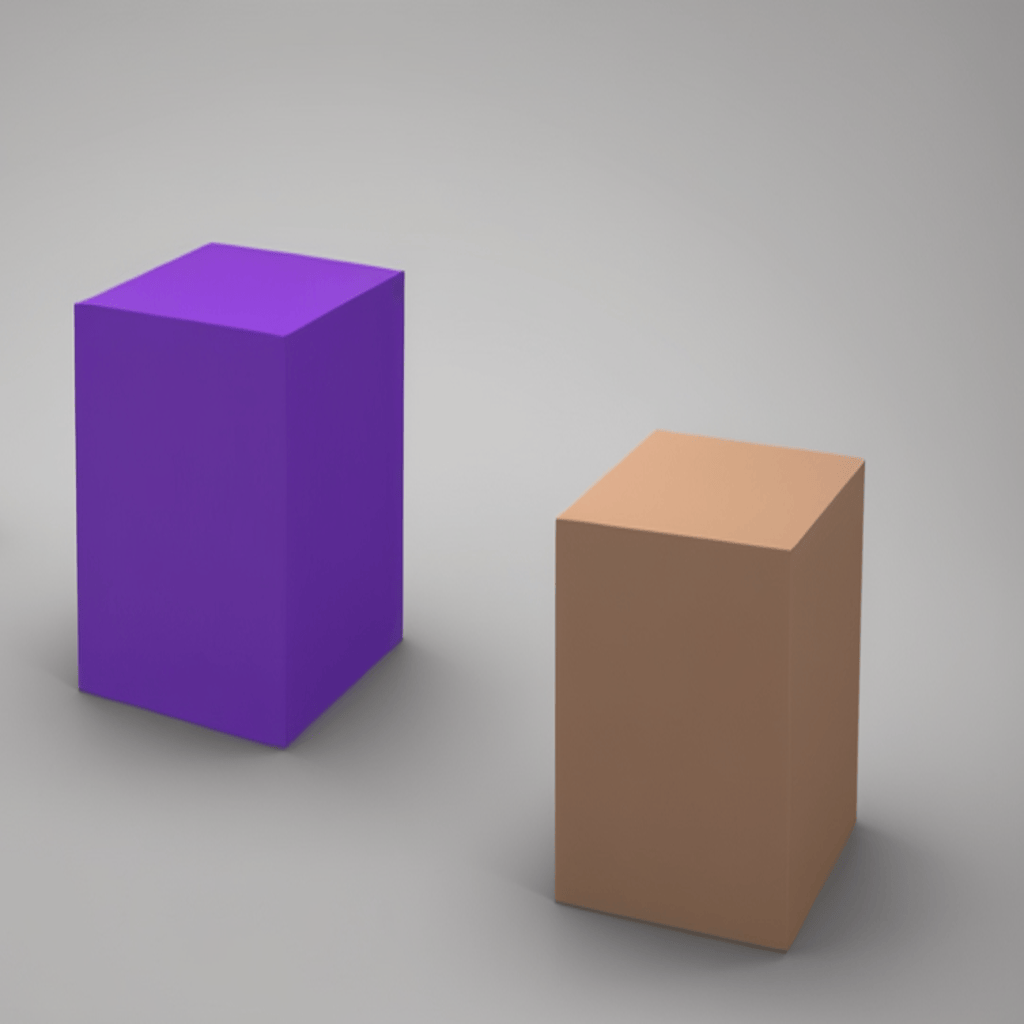} \\
        \small (a) Plain prompt & \small (b) Prompt + Structured information
    \end{tabular}
    \caption{Qualitative comparison on the prompt: ``Add a purple cube at the center. Add a brown cube in front of it on the right.'' (a) SDXL with plain prompt fails to render the intended arrangement. (b) With structured information, the output correctly places a brown cube in front-right of the purple cube.}
    \label{fig:qualitative}
\end{figure}

Figure~\ref{fig:qualitative} presents a qualitative comparison between plain prompts and prompts augmented with structured information. While the plain prompt produces left image with mixed attributes and incorrect spatial arrangements, the addition of structured information resolves these issues and generates the intended scene. These results visually demonstrate the effectiveness of our method for text-to-image generation. 
\subsection{Comparison With Manual Extraction}
As described in Section~\ref{sec:31}, we constructed 100 prompt samples paired with manually extracted structured information. We also used a fine-tuned T5-small model to automatically extract structured information. Both types of structured information were appended to the original plain prompts and used as inputs to SDXL. The empirical results in Table 3 indicate that automatically generated tuples achieve results comparable to manually created tuples, with even slightly better performance on spatial relations. This finding confirms that a lightweight T5-small model is sufficient for reliable tuple generation.

\section{Conclusion}
\label{sec:conclusion}
We proposed a lightweight and portable method to enhance spatial faithfulness in text-to-image generation by appending structured information to plain text prompts. Leveraging a fine-tuned T5-small model with Stable Diffusion XL, our method achieves measurable improvements in spatial accuracy, as validated by Qwen2.5-VL judgments and the Inception Score. The approach is plug-and-play, requiring no retraining of diffusion models, introduces minimal computational overhead since a small language model suffices, and delivers strong improvements in spatial reasoning while preserving visual quality.

\vfill\pagebreak

\bibliographystyle{IEEEbib}
\bibliography{strings}

\end{document}